%% file: root.tex
\documentclass[letterpaper, 10 pt, conference]{ieeeconf}  % Comment this line out if you need a4paper

\IEEEoverridecommandlockouts                              % This command is only needed if 
\usepackage{amsmath} % assumes amsmath package installed
\usepackage{amssymb}  % assumes amsmath package installed
\usepackage{subfiles}
\usepackage{graphicx} 
\usepackage{balance}
\usepackage{comment}

\input{macro.tex}

\title{\LARGE \bf
Differentiable Simulation of Soft Robots with Frictional Contacts
}

\author{Author Names Omitted for Anonymous Review}

\author{Etienne Ménager$^{1}$, Louis Montaut$^{1,2}$, Quentin Le Lidec$^{1}$ and Justin Carpentier$^{1}$% <-this % stops a space
\thanks{$^{1}$  Inria and Département d’Informatique de l’École Normale Supérieure, PSL Research University in Paris, 75013 Paris, France.
$^{2}$ Czech Institute of Informatics, Robotics and Cybernetics, Czech Technical University in Prague, 16000 Praha, Czech Republic.
France (e-mail: etienne.menager@inria.fr, louis.montaut@inria.fr, quentin.lelidec@inria.fr, justin.carpentier@inria.fr).}% <-this % stops a space
}

\begin{document}

\maketitle
\thispagestyle{empty}
\pagestyle{empty}

%%%%%%%%%%%%%%%%%%%%%%%%%%%%%%%%%%%%%%%%%%%%%%%%%%%%%%%%%%%%%%%%%%%%%%%%%%%%%%%%
\input{abstract}

%%%%%%%%%%%%%%%%%%%%%%%%%%%%%%%%%%%%%%%%%%%%%%%%%%%%%%%%%%%%%%%%%%%%%%%%%%%%%%%%
\subfile{1-intro}
\subfile{2-background}
\subfile{3-method}

\subfile{4-exp}

\input{conclusion}


%%%%%%%%%%%%%%%%%%%%%%%%%%%%%%%%%%%%%%%%%%%%%%%%%%%%%%%%%%%%%%%%%%%%%%%%%%%%%%%%

%%%%%%%%%%%%%%%%%%%%%%%%%%%%%%%%%%%%%%%%%%%%%%%%%%%%%%%%%%%%%%%%%%%%%%%%%%%%%%%%

%%%%%%%%%%%%%%%%%%%%%%%%%%%%%%%%%%%%%%%%%%%%%%%%%%%%%%%%%%%%%%%%%%%%%%%%%%%%%%%%

\section*{ACKNOWLEDGMENT}

This work was supported by a grant overseen by the French National Research Agency (ANR) and France 2030 as part of the Organic Robotics Program (PEPR O2R) and the PR[AI]RIE-PSAI AI cluster (ANR-23-IACL-0008), by the French government under the management of Agence Nationale de la Recherche through the project NIMBLE (ANR-22-CE33-0008), and by the  European Union through the AGIMUS project (GA no.101070165).

%%%%%%%%%%%%%%%%%%%%%%%%%%%%%%%%%%%%%%%%%%%%%%%%%%%%%%%%%%%%%%%%%%%%%%%%%%%%%%%%

\bibliography{bibliography}
\bibliographystyle{ieeetr}
\balance{}

\end{document}

%% file: macro.tex
\usepackage{xcolor}

%% file: abstract.tex
\begin{abstract}
    In recent years, soft robotics simulators have evolved to offer various functionalities, including the simulation of different material types (e.g., elastic, hyper-elastic) and actuation methods (e.g., pneumatic, cable-driven, servo-motor). These simulators also provide tools for various tasks, such as calibration, design, and control. However, efficiently and accurately computing derivatives within these simulators remains a challenge, particularly in the presence of physical contact interactions.
    Incorporating these derivatives can, for instance, significantly improve the convergence speed of control methods like reinforcement learning and trajectory optimization, enable gradient-based techniques for design, or facilitate end-to-end machine-learning approaches for model reduction. 
    This paper addresses these challenges by introducing a unified method for computing the derivatives of mechanical equations within the finite element method framework, including contact interactions modeled as a nonlinear complementarity problem. 
    The proposed approach handles both collision and friction phases, accounts for their nonsmooth dynamics, and leverages the sparsity introduced by mesh-based models. Its effectiveness is demonstrated through several examples of controlling and calibrating soft systems.
    
\end{abstract}

%% file: 1-intro.tex
\section{Introduction}
\label{sec:intro}

Soft robotics is a research field that offers unprecedented flexibility, adaptability, and safety in various applications ranging from medical devices to search and rescue operations~\cite{articleLaschiOpportunities}. Compared to traditional rigid robots, soft robots can deform and adapt to their environment, mimicking the capabilities of biological organisms. However, the design and control of soft robots present significant challenges due to their complex dynamics and interactions with the environment~\cite{articleRusChallenges}. Physics-based simulation of soft robots provides a powerful tool to address these challenges~\cite{articleChoiOnTheUseOfSimu}, offering a safe, fast, and cost-effective approach to test designs and control strategies. 

Soft robots simulation involves three key elements: collision detection, resolving contact forces based on physical laws, and computing the robot’s response to these forces and actuation inputs. For simulations to be effective, they must accurately capture the robot’s behavior and its interactions with the environment, including nonsmooth collision detection and resolution. This requires mechanical models capable of handling various types of actuation, geometry, behavior laws, and realistic contact interactions. Additionally, simulators can leverage model-derived information, such as derivatives. Using derivatives information in soft robotics simulation could improve the efficiency of control methods like reinforcement learning and trajectory optimization through first-order information, support gradient-based design techniques, or enable end-to-end machine learning approaches in soft robotics.

Over the past few years, several open-source soft robot simulators have been developed within the community, offering subsets of these features. Beam-based simulators, such as SoroSim~\cite{articleSoroSym} and Elastica~\cite{articleElastica}, have been used for control tasks but are limited to simulating slender structures. For more general geometries, 3D Finite Element Method (FEM) simulators like SOFA~\cite{Faure2011SOFAAM} and Sorotoki~\cite{articleSorotoki} allow simulating a wide range of material models and actuation types for control and design tasks. However, none of these simulators support full differentiation of the simulation pipeline, and their contact models are often simplified, typically relying on the Signorini law or linearized friction cones. Some simulators do offer full pipeline differentiation. For example, DiffTachi~\cite{Hu2019DiffTaichiDP}, ChainQueen~\cite{Hu2018ChainQueenAR}, and SoMoGym~\cite{articleSomogym} use the Material Point Method to simulate soft robots and provide a differentiable programming environment that allows for gradient-based strategy learning. However, these simulators cover only a limited range of soft robots and simplified contact equations. Self-collisions are modeled using momentum transfer between neighboring soft body particles, while collisions between soft bodies and rigid body obstacles are treated as boundary conditions. Position-based methods typically handle collision detection. DiffPD~\cite{Du2021DiffPDDP} offers a fully differentiable FEM simulation of soft robots but uses a linearized contact model. More recently, differentiable simulation for pneumatic actuation has been implemented within the FEM simulation framework~\cite{penaumticDifferentiable}. This simulator focuses on optimizing pneumatic actuation designs, and contact modeling is performed using incremental potential contact methods~\cite{Huang2022DifferentiableSF}, which rely on unsigned distance functions and barrier methods but do not solve the full nonlinear complementarity problem (NCP) involved in contact modeling.

Despite the progress made in differentiable simulations for soft robotics, no solution fully captures the nonsmooth interactions between soft robots and their environments, particularly regarding collision detection and frictional contact dynamics. In this paper, we introduce a feature-complete differentiable simulation pipeline that accounts for soft robot FEM modeling, NCP-based contact formulation, and variation of contact geometries, leveraging and extending recent works in the field of differentiable simulation of poly-articulated rigid robots~\cite{montaut2022differentiablecollisiondetectionrandomized,lidec2024endtoendhighlyefficientdifferentiablesimulation}. 
 In particular, we show how to differentiate the continuum mechanics equations and apply implicit differentiation to compute the NCP gradient for frictional contact problems in soft systems. Additionally, we extend differentiable collision detection, initially developed for rigid robots, to handle deformable objects. We illustrate the effectiveness of these methods through several examples, including the identification of mechanical parameters and the solution of inverse dynamics, both with and without contact. To facilitate further research and ensure reproducibility, the code is publicly available.

The paper is organized as follows. In Sec.~\ref{sec:background}, we provide an overview of soft robotics simulation, covering FEM modeling and contact interactions. Sec.~\ref{sec:method} presents the core contribution of this paper, where we detail how the gradients of the FEM modeling, collision detection, and contact forces are combined to achieve end-to-end differentiability. In Sec.~\ref{sec:exp}, we leverage our differentiable physics simulator to tackle various optimization and control problems, both with and without contact, using different soft systems, including a deformable beam, a Trunk robot, soft Fingers, and a soft Gripper. Finally, Sec.~\ref{sec:conclusion} discusses the limitations of this work and outlines potential future directions.

%% file: 2-background.tex
\section{Background}
\label{sec:background}

In FEM simulation, a soft robotic system is typically discretized and represented by a mesh composed of volumetric elements. This discretization serves as the foundation of the simulation, enabling a detailed representation of both the robot's internal mechanics and its interactions with the surrounding environment. 
In this section, we first recall the FEM modelling of the robot. We then discuss the contact interactions, constitutive laws, and actuation methods employed for simulation.

\subsection{FEM modelling in soft robotics}
\label{subsec:FEMmodelling}

The mechanical behaviour of soft robots is described using a continuum mechanics equation, for which there are no analytical solutions in the general case. Non-linear FEM is one of the numerical methods used to compute a converging approximate solution. Using an implicit time integration scheme and a first-order linearisation of the internal forces, the equation of dynamics is written: 
\begin{align}
    A(x_i, p) &= M + h D(x_i, p) + h^2 K(x_i, p) \nonumber \\
    b(x_i, v_i, p) &= -h^2 K(x_i, p) v_i + h(P - f_i(x_i, p) ) \label{eq:FEMmodelling}\\
    A(x_i, p) dv &= b(x_i, v_i, p) + hH_a(x_i)^T \lambda_a + hH_c(x_i)^T \lambda_c \nonumber,
    % \label{eq:FEMmodelling}
\end{align}
where $(x_i, v_i)$ are the position and velocity of the nodes of the mesh, $p$ are material parameters, $A$ is the impedance matrix, $M$ the mass matrix, $D = \alpha M + \beta K$ is the damping matrix defined using Rayleigh damping, $K$ is the stiffness matrix, $dv = v_f -v_i$ the difference between final speed and initial speed, $P$ the external forces, $f_i$ the internal forces, $H_a^T\lambda_a$ the actuation forces, $H_c^T\lambda_c$ the contact forces and $h$ the time step. The dependence of quantities on $x_i$, $v_i$, and $p$ is highlighted. 

This system of equations is solved in two stages. First, the free configuration of the robot is computed by solving the linear system of Eq.~\eqref{eq:FEMmodelling}, using a sparse linear solver. This is the state of the robot when the contact forces are zero. This configuration is then corrected using the value of the contact forces $\lambda_c$, which are calculated as a solution to the associated NCP (see Sec.~\ref{subsec:contactModelling}). The positions of the mesh points are then updated using the semi-explicit scheme $x_f = x_i + h v_f$.

\subsection{Interactions with the external environment}
\label{subsec:contactModelling}

In soft robotics, interactions with the environment that do not involve actuation are primarily achieved through contact. Resolving these contacts involves two key steps: first, collision detection, which identifies the contact points and contact normal, enabling the construction of the contact Jacobian $H_c$; and second, collision resolution, which calculates the contact forces, denoted as $\lambda_c$.

\noindent\textbf{Collision detection.}  
This phase determines the contact information between two colliding geometries. For a given pair of bodies, a collision detection algorithm computes a contact point and a contact normal, representing the direction separating the bodies with minimal displacement. In this article, we utilize a collision detection algorithm based on the GJK method~\cite{Montaut2022CollisionDA}. The matrix $H_c$, which defines the directions of the contact forces, is populated with contact frames $c$. Each frame is centered at the contact point, with its $z$-axis aligned along the contact normal.

\noindent\textbf{Collision resolution.} Contact modelling involves various complementary physical principles~\cite{contactLidec}. First, the so-called Signorini condition provides a complementarity constraint $ 0 \leq \lambda_{c,N} \perp \sigma_{c,N} \geq 0$, where $\sigma_c$ is the velocity of the contact points and $_N$ stands for the normal component of the vector. The Signorini condition ensures that the normal force is repulsive, the bodies do not interpenetrate, and the power injected by normal contact forces is null. Secondly, the maximum dissipation principle combined with the frictional Coulomb law $\|\lambda_{c,T}\| \leq \mu \lambda_{c,N}$ of friction $\mu$, states that the tangential component of the contact forces $\lambda_{c, T}$ maximizes the power dissipated by the contact. 
These principles are equivalent to an NCP: 
\begin{align}
     \mathcal{K}_\mu &\ni \lambda_c \perp \sigma_{c} + \Gamma_\mu(\sigma_c) \in \mathcal{K}_\mu^* \label{eq:contact} \\
    \sigma_{c} &= G \lambda_c + g \nonumber
\end{align}
where $G$ is the Delassus matrix that gives the admittance matrix $A^{-1}$ projected on the contacts and enables to map contact forces to contact points velocities, $g$ is the free velocity of the contact points when $\lambda_c = 0$, $\mathcal{K}_{\mu}$ is a second-order cone with aperture angle $\text{atan}(\mu)$, $\mathcal{K}_{\mu}^*$ is the dual cone of $\mathcal{K}_{\mu}$, and $\Gamma_\mu(\sigma_c) = [0, 0, \mu \| \sigma_{c,T}\|]$ is the De Saxcé correction enforcing the Signorini condition~\cite{de1998bipotential,acary2018solving,contactLidec}. This NCP can be solved to find the value of $\lambda_c$ using different optimization methods, like Projected Gauss-Seidel methods (PGS)~\cite{acary2018solving,contactLidec} or the Alternating Direction Method of Multipliers (ADMM)~\cite{acary2018solving,Carpentier2024FromCT}.

\subsection{Soft Robot's internal mechanics}
\label{subsec:behaviorModelling}

The internal mechanical behaviour of soft robots is modelled using two key elements. First, the internal forces $f_i$ of the soft robot are derived from the constitutive laws. To form the linearized system~\eqref{eq:FEMmodelling}, the derivatives of these internal forces, $K = \frac{\partial f}{\partial x}$ and $D = \frac{\partial f}{\partial v}$, are calculated. Second, actuators control the system's deformation by applying a force $\lambda_a$. The effects of this actuation are modelled in the Jacobian of the actuation, denoted as $H_a$.

\noindent\textbf{Constitutive laws.} Constitutive laws define the relationship between stress, strain, and other material properties. In this paper, we consider one elastic and two hyper-elastic constitutive laws: co-rotational elasticity, Saint-Venant-Kirchhoff, and Neo-Hookean. They are often described using a strain energy density function $\Psi$, that can be computed using mechanical quantities such as the deformation gradient $F$, the Green-Lagrangian strain tensor $E$, and/or the right Cauchy-Green tensor $C$, evaluated in each volumetric element of the discretization. These energies are used to calculate the internal forces and the stiffness matrix, using spatial discretisation, the Nanson’s formula and the internal virtual work. A comprehensive overview of these constitutive laws and associated computations is given in~\cite{phdthesisBrunet}. For tetrahedron discretization, the final expression of the internal forces is given by:
\begin{equation}
    f_{e,i} = F_e S_e \cdot \nabla N_i
    \label{eq:internalforces}
\end{equation}
where $_e$ stands for the $e$th tetrahedron in the mesh, $i$ the $i$th node of the tetrahedron, $S_e = 2 \frac{\partial \Psi}{\partial C}$ the second Piola-Kirchhoff stress tensor, and $\nabla N_i$ the derivative of the shape function in the global reference. 

\noindent\textbf{Actuation modelling.} 
The actuators are modelled using the Jacobian matrix $H_a$, which links the force $\lambda_a$ to the position and speed of the nodes in the mesh. The matrix $H_a$ corresponds to the direction of the actuation constraint~\cite{articleActuation}. In this paper, we consider three kinds of actuation: cable actuation for which $H_a$ is filled with the direction of the cable; pneumatic actuation for which $H_a$ is filled with the oriented area of the surface of the corresponding cavity; and servomotors for which $H_a$ is filled with $1$ in the direction of the servomotor actuation.

%% file: 3-method.tex
\section{Differentiable Soft-Robotics Simulation with Contact NCP}
\label{sec:method}

This section details the core contribution of this paper, namely a differentiable soft robotics simulator that combines differentiable collision detection for deformable body, differentiable contact resolution and differentiable FEM modelling of soft robot. We develop these three points in the remainder of this section. We notably extend the work presented in~\cite{lidec2024endtoendhighlyefficientdifferentiablesimulation} and~\cite{montaut2022differentiablecollisiondetectionrandomized} tailored for rigid poly-articulated systems (e.g., humanoids, manipulators, etc.) to the context of soft robotics. 

In the following, let $\theta$ be the differentiation variable, which is any subset of the inputs $\{x_i, v_i, \lambda_a\}$ or physical parameters $p$ like the Young Modulus of the material and $\lambda_c^*$ the solution of the NCP problem (Eq.~\ref{eq:contact}).

\subsection{Collision detection derivatives}
\label{subsec:collisionderivatives}

The collision detection stage depends on the position of the tetrahedron in the mesh, and thus of the position $x_i$. This can be highlighted by explicitly stating the dependency $H_c(x_i)$. When $\theta$ depends on the position, the terms $\partial_\theta H_c^T$ and $\partial_\theta H_c$ are non-null. Calculating these terms means differentiating the collision detection stage. The general idea of the derivation of the collision detection is as follows, and uses the various properties of the collision detection algorithm.

Let $T_1$ and $T_2$ be two colliding tetrahedra. Let \mbox{$\mathcal{D} = \{p_1 - p_2 : p_1 \in T_1, p_2 \in T_2 \}$} be the Minkowski difference between $T_1$ and $T_2$. The GJK algorithm solves the detection collision problem as a min-norm-point optimization problem:
\begin{equation}
    \min_{x \in \mathbb{D}} \frac{1}{2}\| x\| ^2 
    \label{eq:GJK}
\end{equation}
by calculating a sequence of support functions \mbox{$\partial \sigma_{T_i}(d) = \arg \min_{x \in T_i} x^T d$} in the direction $d = \nabla \| x\| ^2$ for each tetrahedron. 
Importantly, each support function depends on the position of the points composing the tetrahedron. This is expressed by \mbox{$\partial \sigma_{T_i}(d) = \arg \min (d^T x_{i1}, d^T x_{i2}, d^T x_{i3}, d^T x_{i4})$}, where $x_{ij}$ are the vertices of the tetrahedron. The solution $x^*$ of the optimization problem (\ref{eq:GJK}) is the separator vector of the smallest norm. This vector characterizes the collision in terms of normal and contact points. 
We refer to~\cite{Montaut2022CollisionDA} for more details.

The implicit function theorem can be used to calculate the derivatives of the separator vector with respect to the position of the points on the tetrahedron. By definition,  $x^*$  is the  solution of the equation:
\begin{equation}
    f(x^*, T_1, T_2) = x^* + \partial \sigma_{T_1}(x^*) - \partial \sigma_{T_2}(x^*) \ni 0.
    \label{eq:detectionImplicite}
\end{equation}
From the $1^\text{st}$-order Taylor expansion of $f$, one can relate the sensitivity of the optimal solution $x^*$ to the relative position of the tetrahedron $T_i$:
\begin{equation}
    \frac{\partial f}{\partial x^*} \delta x^* + \frac{\partial f}{\partial T_i}\delta T_i = 0,
\end{equation}
leading to the following relation:
\begin{equation}
    \frac{\partial f}{\partial x^*} \frac{\partial x^*}{\partial T_i} = - \frac{\partial f}{\partial T_i}.
    \label{eq:detectionImpliciteDerivative}
\end{equation}

When several collision points exist between the different shapes, the variation of the normal contact direction can lead to a reduction of the support set to one point. 
In this configuration, the Hessian of the support function is uninformative and sometimes ill-defined. To get an idea of the variation of $f$, it is possible to use a randomized smoothing technique with limited computational overhead~\cite{montaut2022differentiablecollisiondetectionrandomized}. 
In the case of a tetrahedron, using a Gumble distribution~\cite{montaut2022differentiablecollisiondetectionrandomized} leads to a close form of the $\arg\min$ operator. 
This provides an estimate of the derivative of $f$ with respect to $x^*$ and $T_i$, and thus an expression for the derivative of collision detection by solving the complete system.

\subsection{NCP derivatives using implicit differentiation}
\label{subsec:NCPderivatives}

Calculating $\partial_\theta \lambda_c^*$ means differentiating the collision resolution stage. Computing the value of $\frac{\partial \lambda_c^*}{\partial \theta}$ is challenging, as $\lambda_c^*$ is the non-smooth solution of the equation  \mbox{$ \text{NCP}(G, g) = \text{NCP}(\lambda_c^*, x_i, v_i, \lambda_a, p) = 0 $}. In addition, contact can switch between three distinct modes, namely breaking contact, sticking contact, and sliding contact. As proposed in~\cite{lidec2024endtoendhighlyefficientdifferentiablesimulation}, the implicit function theorem can be used to compute $\frac{\partial \lambda_c^*}{\partial \theta}$  by deriving the optimality conditions of the problem, assuming that the contact mode is fixed. We have not yet considered the case of computing derivatives at the boundary between two contact modes. This remains an open issue for the robotics community.

In the case of contact forces, the derivatives can be written as the solution of a linear system:
\begin{equation}
    \bar A X = - \bar B \left(\frac{\partial G}{\partial \theta} \lambda_c^* + \frac{\partial g}{\partial \theta} \right),  \label{eq:implicitdifferentiation}
\end{equation}
where $\bar A$ and $\bar B$ are two matrices construct with $G$, $\sigma_c$ and $\lambda_c$, according to the mode of the contact points (see ~\cite{lidec2024endtoendhighlyefficientdifferentiablesimulation} for more details). Using the fact that $G \lambda_c^* + g = \sigma_c = H_c v_f$ is the contact points' velocity, the derivative $\frac{\partial G}{\partial \theta} \lambda_c^* + \frac{\partial g}{\partial \theta}$ corresponds to the derivative of the contact point velocity with $\lambda_c = \lambda_c^*$ constant. Finally, this derivative can be written: 
\begin{equation}
    \frac{\partial G}{\partial \theta} \lambda_c^* + \frac{\partial g}{\partial \theta} = H_c \frac{\partial v_f}{\partial \theta} \Big|_{\lambda_c = \lambda_c^*} + \frac{\partial H_c \cdot v_f}{\partial \theta}\Big|_{\lambda_c = \lambda_c^*}
\end{equation}
The first term is the solution of Eq.~\ref{eq:derivativevf} with the contact force taken constant. The second term involves the derivative of the contact Jacobian, as computed in the previous section. Finally, the derivatives $\frac{\partial \lambda_c^*}{\partial \theta}$ are obtained by solving Eq.~\ref{eq:implicitdifferentiation} using a QR decomposition of $\bar A$. 

\subsection{Chaining FEM model derivatives and contact derivatives}
\label{subsec:globalderivative}

Starting from Eq.~\ref{eq:FEMmodelling}, we can write the final velocity as a combination of the free velocity $v_f^\text{free}= v_i + A^{-1} b$, actuation correction $\delta_a = hA^{-1}H_a^T \lambda_a$ and contact correction \mbox{$\delta_c = hA^{-1}H_c^T \lambda_c^*$}. The goal is to compute the Jacobian $\frac{\partial v_f}{\partial \theta}$, as the derivatives of $x_f$ can be written $\frac{\partial x_f}{\partial \theta} = \frac{\partial x_i}{\partial \theta} + h \frac{\partial v_f}{\partial \theta}$.
Differentiating $v_f$ leads to: 
\begin{equation}
    \frac{\partial v_f}{\partial \theta} =  \frac{\partial v_f^\text{free}}{\partial \theta} + \frac{\partial \delta_a}{\partial \theta} +  \frac{\partial \delta_c}{\partial \theta}, 
    \label{eq:derivativevf}
\end{equation}
where all terms can be expanded using the chain rules.

The derivatives $\frac{\partial v_f^\text{free}}{\partial \theta}$ and $\frac{\partial \delta_a}{\partial \theta}$ can be obtained using FEM modelling (Eq.~\ref{eq:FEMmodelling}), as all elements ($A$, $b$, $K$, $D$) can be expressed with the internal forces (Eq.~\ref{eq:internalforces}) and its derivatives and the computation of the Jacobian matrix $H_a$ (Sec.~\ref{subsec:behaviorModelling}). FEM modelling shows that the $H_a$ and $K$ matrices are always multiplied by vectors. This means that it is not useful to calculate $\partial_\theta H_a$ or $\partial_\theta H$ (higher order tensor) but $\frac{\partial H_a \cdot V}{\partial \theta}$ and $\frac{\partial K \cdot V}{\partial \theta}$ where $V$ is any vector. Considering tetrahedron elements with 4 nodes $[P_1, P_2, P_3, P_4]$, the computation of the derivatives for a tetrahedron can be expressed as a matrix in $\mathbb{R}^{n_v \times 12}$ where $n_v$ is the dimension of the initial vector. The global derivative is a combination of the derivatives calculated at the level of each tetrahedron, arranged according to the nodes contained in the tetrahedron. 

The derivative $\frac{\partial \delta_c}{\partial \theta}$ is the sum of three elements: 
\begin{equation}
    h \frac{\partial A^{-1} \cdot (H_c^T \lambda_c^*)}{\partial \theta} +  h A^{-1} \frac{\partial H_c^T \cdot \lambda_c^*}{\partial \theta} +  h A^{-1} H_c^T \frac{\partial \lambda_c^*}{\partial \theta}
    \label{eq:ddeltac}
\end{equation}
The value of $H_c$ and $\lambda_c^*$ are computed using collision detection algorithm and collision resolution algorithm. The first term can be computed using the derivative of $K$ as explained in the previous paragrapher, and the fact that $\partial_\theta A^{-1} = - A^{-1} \partial_\theta A A^{-1}$, resulting from $AA^{-1} = \text{Id}$. The second term involves the derivative of the Jacobian $H_c$ (collision detection), discussed in section \ref{subsec:collisionderivatives}. The third term involves the derivative of $\lambda_c^*$ (collision resolution), discussed in section \ref{subsec:NCPderivatives}.

%% file: 4-exp.tex
\section{Experiments}
\label{sec:exp}

In this section, we apply our approach to address optimization and control tasks across various soft systems. We begin by using simple toy examples to validate the computation of derivatives. Next, we describe several soft systems along with their corresponding tasks. Finally, we present results for both contact and non-contact scenarios. 

\noindent\textbf{Implementation details.} The FEM modelling of the soft systems and actuation are based on modified model developed in SOFA framework~\cite{Faure2011SOFAAM}~\cite{phdthesisBrunet}. The analytical derivatives are computed in Python, with future developments in C++ for efficiency. The collision detection is realized using HPP-FCL~\cite{hppfcl} for efficient collision detection. The contact resolution is achieved using ADMM-based approach~\cite{Carpentier2024FromCT} with sparse linear backend leveraging. The code associated with this paper will be released as open-source. All the experiments are performed on a single core of an Apple M3 CPU.

\begin{figure}[!ht]
\centering
\resizebox{0.4\textwidth}{!}{
\includegraphics{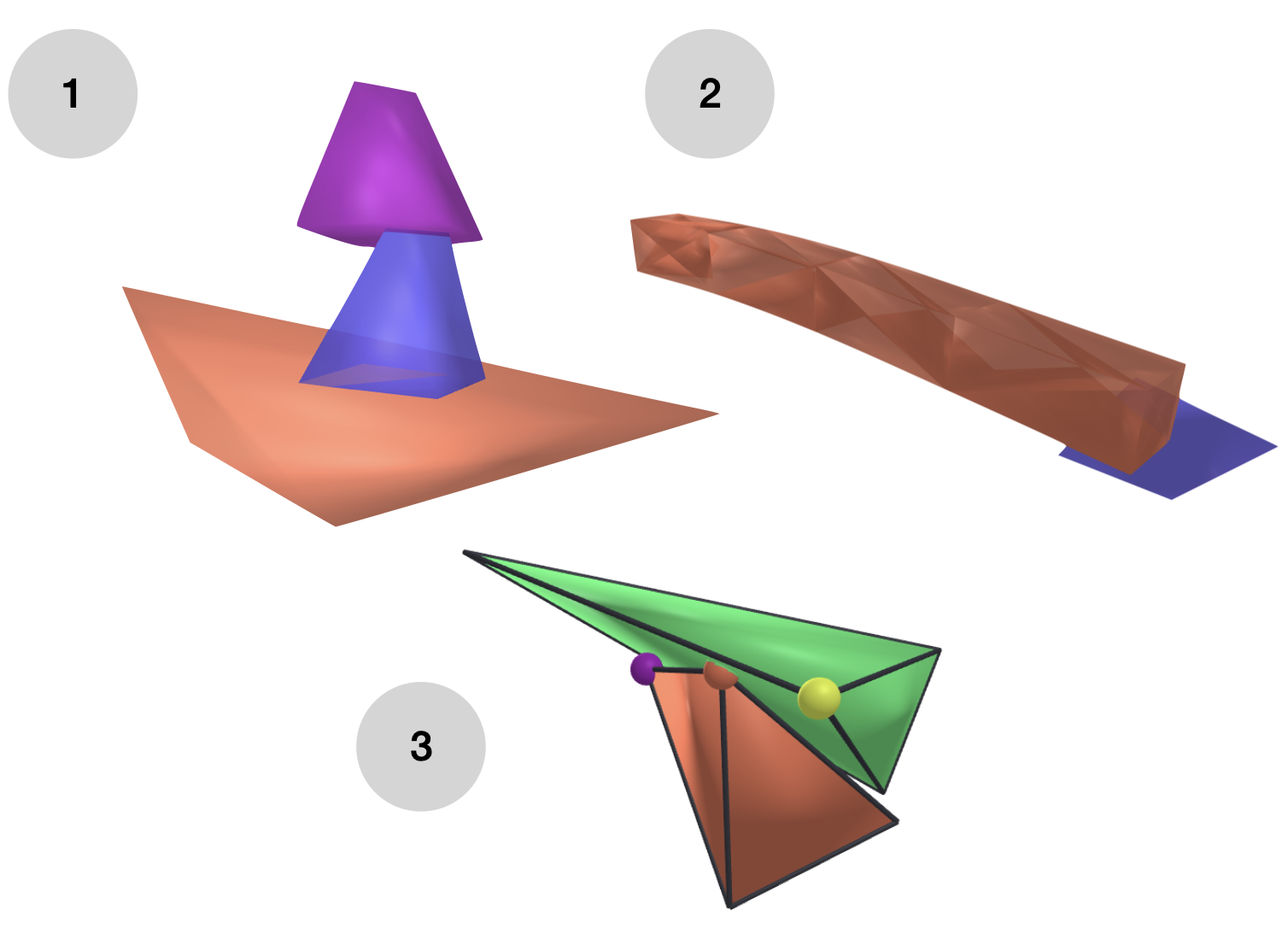}
}
\caption{Toy examples developed to test the calculation of derivatives. (1) Multi-contact and multi-material. (2) Actuation, constrained motion, and rigidification. (3) Collision detection and optimization of contact point positions.  }
\label{fig:toyexamples}
\end{figure}

\subsection{Verification of analytical derivatives}
\label{subsec:toyexample}

\begin{figure*}[!ht]
\centering
\resizebox{0.9\textwidth}{!}{
\includegraphics{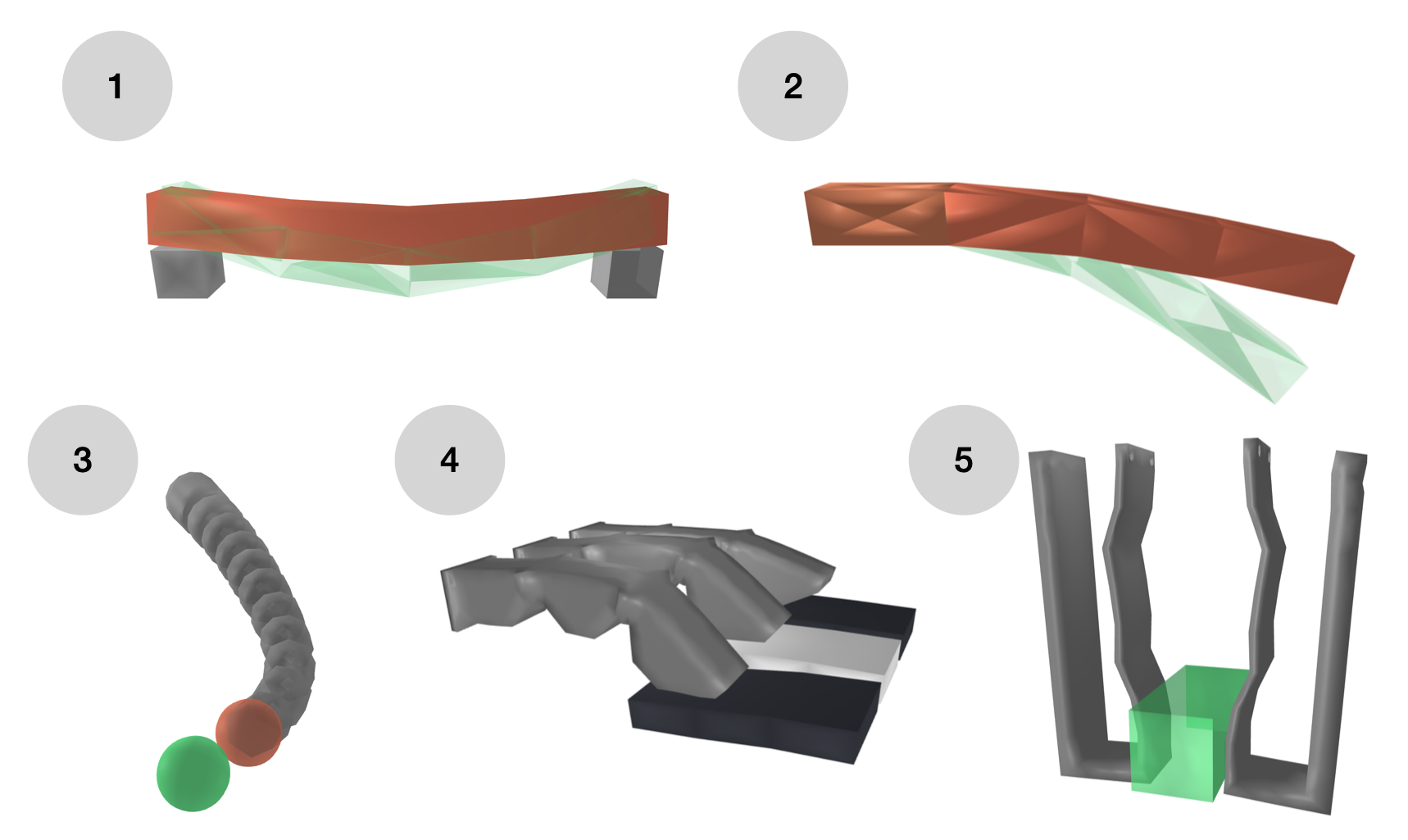}
}
\caption{Simulated deformable systems considered in this work. (1) Deformable beam with contact. (2) Deformable beam without contact. (3) Trunk robot. (4) The robot pianist, composed of three identical Finger robots. (5) Soft Gripper. }
\label{fig:sofsystem}
\end{figure*}

To validate the derivative calculations introduced in the previous section, we present a series of toy examples, some of which are illustrated in Fig.~\ref{fig:toyexamples}. When the finite-difference method is applicable — i.e., when no switching occurs between contact modes, as it is possible with position derivatives — the derivatives are compared against the finite-difference method. The results show a maximum error on the order of $10^{-4}$, which is within expected bounds. This method is used to numerically verify derivatives with respect to $x_i$, $v_i$, $\lambda_a$ and $p$, in scenarios involving multiple objects and constitutive laws (Fig.~\ref{fig:toyexamples}.1), as well as in cases with additional constraints such as rigidification (where some degrees of freedom are treated as rigid), projective constraints (where some degrees of freedom are fixed), or actuation (Fig.~\ref{fig:toyexamples}.2). These tests are conducted both with and without contacts. In cases where the finite-difference method is not valid for derivative evaluation, the computed gradients are used to perform specific optimization tasks. For example, collision detection differentiation is tested by optimizing the position of a point on a tetrahedron to become the point of contact (Fig.~\ref{fig:toyexamples}.3). These various examples help to verify the accuracy and behaviour of the computed gradients.

\subsection{Soft-systems and associated tasks}
\label{subsec:softsystems}

We proposed to illustrate our approach with five different systems, presented in Fig.~\ref{fig:sofsystem}:

\begin{enumerate}
    \item The deformable beam consists of a deformable body that bends under its own weight. In the contact scenario (see Fig~\ref{fig:sofsystem}.1), the beam rests on two supports. In the non-contact scenario (see Fig.~\ref{fig:sofsystem}.2), the beam is attached at one end. These examples are used to identify the mechanical parameters of the system through an optimization method applicable in both contact and non-contact situations.
    \item The Trunk robot~\cite{Coevoet2017OptimizationBasedIM} (see Fig.~\ref{fig:sofsystem}.3) is a soft robot composed of a flexible body made out of silicone and actuated by 8 cables fixed all along the robot. This robot is used for inverse dynamics tasks.
    \item The pianist robot (see Fig.~\ref{fig:sofsystem}.4) is composed of three identical soft Finger robots. Each Finger robot is made up of three segments connected by accordion-shaped joints and actuated by 2 cables, as proposed in~\cite{Navarro2020AMS}. The robot fingers interact with piano keys, which can tilt according to the effort exerted on them. This robot is used for inverse dynamics tasks with contact handling. 
    \item The Soft Gripper (see Fig.~\ref{fig:sofsystem}.3) is composed of two deformable Fingers and one object to be grasped. Each deformable finger is made up of flexible filament and is actuated by a servomotor. These Fingers are presented in~\cite{navez2024design} and are designed to interact with the environment through self-contact and external contact. This robot is used in an optimisation task involving both actuation and contact forces.
\end{enumerate}

\subsection{Identification of material properties with the deformable beam}
\label{subsec:materialOptim}

\begin{figure*}[!ht]
\centering
\resizebox{0.9\textwidth}{!}{
\includegraphics{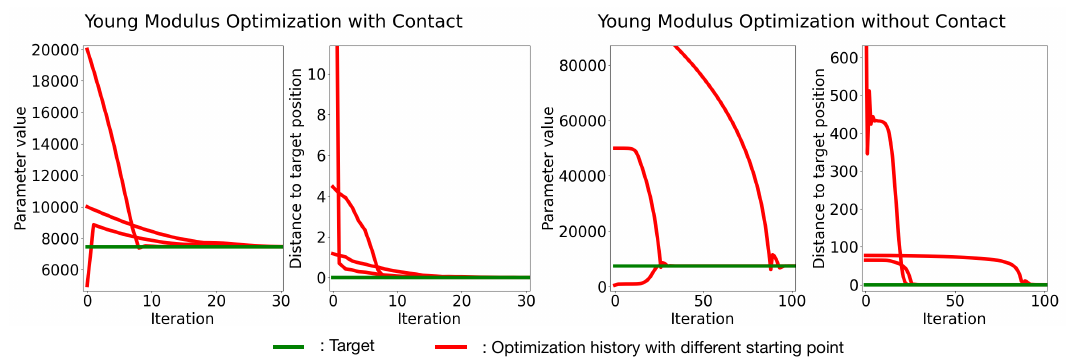}
}
\caption{Optimization of the Young Modulus of the beam to achieve a target position for the mesh nodes. Young's modulus is expressed in MPa, distance in mm. Only the useful parts of the graphs are retained. }
\label{fig:resOptimParam}

\resizebox{0.9\textwidth}{!}{
\includegraphics{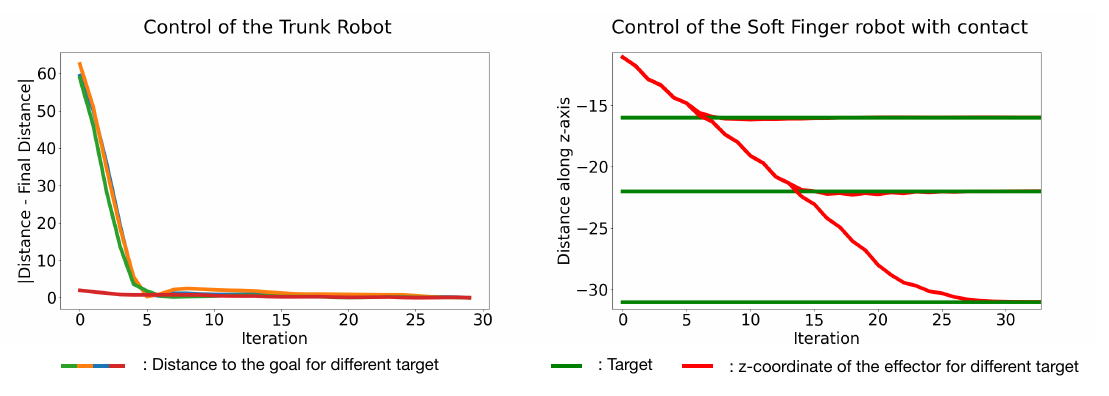}
}
\caption{Evolution of the distance to the target for control tasks using the Trunk robot and the Finger robot. For the Trunk robot, the current position is compared to the final position, as the target may lie outside the robot's workspace. Distances are measured in mm.}
\label{fig:resControl}

\end{figure*}

In this experiment, we assume that the state of a beam subjected to gravity is known after 0.3 seconds, i.e., $x_{t=0.3}(E^*, \mu^*)$ with $E$ Young's modulus, $\mu$ Poisson's ratio, and $^*$ stands for the required parameter value. We therefore define the objective to be minimised as: 
\begin{equation}
    \min_p \frac{1}{2}\| x_{t=0.3}(p) - x_{t=0.3}(E^*, \mu^*) \|^2,
    \label{eq:costfunctionbeam}
\end{equation}
where $p\in \{E, \mu \}$ is the parameter to be optimized (the other one being fixed at its optimum value). This problem is solved using the Levenberg-Marquardt algorithm~\cite{More1977TheLA} with heuristic scheduling on the method's damping parameter $\lambda_{LM}$. This method uses the value of the Jacobian $J = \frac{\partial x}{\partial p} \in \mathbb{R}^{3n}$  according to: 
\begin{equation}
    p_{k+1} = p_k + \frac{1}{\lambda_{LM}(1 + J^TJ)}J^T(x(p_k) - x(E^*, \mu^*)),
\end{equation}
where $n$ is the number of nodes in the mesh, and $k$ is the index of the iterative process. Note that we have removed the notation $t=0.3$ for better readability. In the case of contact scenario, the Jacobian matrix $J = \frac{\partial x}{\partial p}$ takes into account contact information through the correction term:
\begin{equation}
    \frac{\partial (A^{-1}H_c^T \lambda_c)}{\partial p} = \frac{\partial A^{-1} \cdot (H_c^T \lambda_c)}{\partial p} + A^{-1} H_c^T \frac{\partial \lambda_c}{\partial p}.
\end{equation}

The results are shown in Fig.~\ref{fig:resOptimParam} for Young Modulus identification. Similar results can be obtained for Poisson ratio identification. We can see from these figures that the gradient information relating to the mechanical parameters of the materials can be used to identify the model parameters, even if there are contacts between the beam and some obstacles. It should be noted that in the case of contacts, some initializations are not suitable, such as the one where contacts are broken. In this case, many local minimums exist, and the algorithm may not find the solution. 

\subsection{Control tasks with soft robots}
\label{subsec:materialOptim}

\noindent\textbf{Inverse dynamics.} 
The Trunk robot and the Finger robot can be controlled by minimising the distance between the end-effector of the robot and a goal to be reached. At one time step, this distance depends on the position of the nodes, corrected by the effect of the actuation. 
This leads to the formulation: 
\begin{equation}
    \min_{\Delta \lambda_a} \frac{1}{2}\| X_e(x+\Delta x, \lambda_a + \Delta \lambda_a) - X_\text{goal} \|^2
    \label{eq:controltrunk}
\end{equation}
where $X_e$ is the position of the end-effector and $X_\text{goal}$ the position of the goal. Writing the first-order expansion of \mbox{$X_{e, f} = X_e(x+\Delta x, \lambda_a + \Delta \lambda_a)$} we have:
\begin{equation}
    X_{e, f} = X_e(x, \lambda_a) + \frac{\partial X_e}{\partial x} \Delta x + \frac{\partial X_e}{\partial \lambda_a} \Delta \lambda_a 
\end{equation}

By integrating this expression into Eq.~\eqref{eq:controltrunk} and adding actuation limits to take account of the robot's physical limitations, we obtain a QP problem that can be solved with a suitable solver. In this paper, we use ProxQP~\cite{bambade2023proxqp}.  $\Delta x$  is calculated with respect to the free movement of the robot. In the case of the Finger robot, only the vertical direction is controlled. The different results for both robots are shown in Fig.~\ref{fig:resControl}.

In the case of the Trunk robot, the formulation with gradients leads to the same formulation as the one proposed in the reference articles on soft robotics control using QP approaches~\cite{Coevoet2017OptimizationBasedIM}. The system is over-actuated, leading to an oscillation of the end-effector position. It is usually possible to reduce this oscillation by adding corrective terms into the cost function (on the system energy or the actuation magnitude) or by setting a low-pass filter on the actuation force update. In the case of the Finger robot, the derivative with respect to $\lambda_a$ accounts for contact information, requiring the applied force to be greater in order to counterbalance the contact force. The results show that gradient information with respect to position and actuation can be used to control soft robots. 

\noindent\textbf{Control of contact forces.}
In the case of the gripper, we can define an objective based on two terms: one to close the gripper and one to limit the contact forces. As the system is symmetrical with respect to the origin, we can write this problem for one finger and apply the result to both. As for the previous section, this problem can be written as a QP, where the objective is written: 

\begin{equation}
    \min_{\Delta \lambda_a} ~ \alpha \| X_e \|^2 + \beta \| \lambda_{c, \text{nodes}} \|^2,
    \label{eq:controlGripper}
\end{equation}
with $(\alpha, \beta)$ are weight to choose the importance of both terms, $X_e$ is the position of the effector and $\lambda_{c, \text{nodes}}$ stands for the contact force applied on the nodes of the finger surface. Maximal and minimal actuation are set, to fit robot's capability and force minimal motion. Results are shown in Tab.~\ref{tab:controlForce} and compared to the heuristic where constant maximal actuation force is applied to grasp the object. 

\begin{table}[!h]
\centering
   
    \centering
    \begin{tabular}{| l || c | c |  }
        \hline
         &  $\| X_e \|$ (cm) & $\| \lambda_{c, \text{nodes}} \|$ (cN) \\ \hline
         \textit{Heuristic}  & 1.18 & 0.43 \\  \hline
    	 \textit{Optimized}  & 1.20 & 0.03 \\  \hline
    \end{tabular}
\caption{Value of the two components involved in the loss for the Gripper control after 70 iterations. The total loss is calculated as the weighted sum of these two terms.
   }
    \label{tab:controlForce}
\end{table}

As shown in this table, the proposed heuristic applies a constant maximum actuation force to the robot, reducing the distance from the origin. However, this comes at the cost of increased contact force, resulting in a larger loss. In contrast, the proposed control method does not achieve the same reduction in distance from the origin but effectively limits the contact forces on the robot's surface. In the optimized control strategy, the process begins by clamping the object. Once the two terms - distance from the origin and contact force - become equally important, the gripper gradually loosens its grasp, significantly reducing the contact forces while maintaining the distance to the origin.

%% file: conclusion.tex
\section{Discussions and Conclusion} 
\label{sec:conclusion}

This paper introduces an end-to-end differentiable physics pipeline for soft robotics simulation based on the resolution and the implicit differentiation of an NCP for contacts. The simulator leverages state-of-the-art FEM simulation methods to consider several types of constitutive laws, actuation, and geometry. In addition, the definition of contact as an NCP ensures accurate simulation of interactions, which is necessary to prevent unphysical simulation artifacts. This simulator is illustrated with various deformable systems, including cable or servomotor actuation, various geometries, and scenarios with or without contact. Nonetheless, the simulator is designed to be generic and is not limited to these examples.

The limitations of this approach are inherent to solving continuous media mechanics equations using FEM methods. These mathematical systems can be high-dimensional and resource-intensive to solve. 
Additionally, more advanced optimization algorithms for control or identification tasks could be explored. For instance, the presence of contact can lead to local minima. First-order Reinforcement Learning methods could be leveraged to explore the solution space and overcome these local minima efficiently. However, this simulator is a foundation for developing new algorithms in soft robotics. For instance, it could be used to leverage simulation gradients to accelerate the discovery of complex robot movements in contact, to calculate reduced models using learning methods, or to propose calibration techniques based on computer vision networks.

Another drawback is that gradients with respect to position cannot be tested using finite differences due to the non-smooth nature of the contact problem. Despite this, the calculated gradients provide valuable information for problem-solving. Finally, the proposed differentiable simulation of an unrelaxed physics model is a crucial step toward reducing the Sim2Real gap~\cite{hofer2021sim2real}. Future work should exploit this feature to develop new algorithms that facilitate the transfer between simulation and physical prototypes.

In future work, we plan to provide an open-source C++ implementation of the proposed approach, either by extending Pinocchio~\cite{pinocchioweb,carpentier2019pinocchio} to support soft robots or extending SOFA~\cite{Faure2011SOFAAM} to transform it as a fully differentiable simulator.

%% file: root.bbl
\begin{thebibliography}{10}

\bibitem{articleLaschiOpportunities}
C.~Laschi, B.~Mazzolai, and M.~Cianchetti, ``Soft robotics: Technologies and systems pushing the boundaries of robot abilities,'' {\em Science Robotics}, vol.~1, p.~eaah3690, 12 2016.

\bibitem{articleRusChallenges}
D.~Rus and M.~T. Tolley, ``Design, fabrication and control of soft robots,'' {\em Nature}, vol.~521, no.~7553, pp.~467--475, 2015.

\bibitem{articleChoiOnTheUseOfSimu}
H.~Choi, C.~Crump, C.~Duriez, A.~Elmquist, G.~Hager, D.~Han, F.~Hearl, J.~Hodgins, A.~Jain, F.~Leve, C.~Li, F.~Meier, D.~Negrut, L.~Righetti, A.~Rodriguez, J.~Tan, and J.~Trinkle, ``On the use of simulation in robotics: Opportunities, challenges, and suggestions for moving forward,'' {\em Proceedings of the National Academy of Sciences}, vol.~118, p.~e1907856118, 01 2021.

\bibitem{articleSoroSym}
A.~T. Mathew, I.~B. Hmida, C.~Armanini, F.~Boyer, and F.~Renda, ``Sorosim: A matlab toolbox for hybrid rigid–soft robots based on the geometric variable-strain approach,'' {\em IEEE Robotics and Automation Magazine}, vol.~30, no.~3, pp.~106--122, 2023.

\bibitem{articleElastica}
N.~M. Naughton, J.~Sun, A.~Tekinalp, G.~Chowdhary, and M.~Gazzola, ``Elastica: {A} compliant mechanics environment for soft robotic control,'' {\em CoRR}, vol.~abs/2009.08422, 2020.

\bibitem{Faure2011SOFAAM}
F.~Faure, C.~Duriez, H.~Delingette, J.~Allard, B.~Gilles, S.~Marchesseau, H.~Talbot, H.~Courtecuisse, G.~Bousquet, I.~Peterlik, {\em et~al.}, ``Sofa: A multi-model framework for interactive physical simulation,'' {\em Soft tissue biomechanical modeling for computer assisted surgery}, pp.~283--321, 2012.

\bibitem{articleSorotoki}
B.~Caasenbrood, A.~Pogromsky, and H.~Nijmeijer, ``Sorotoki: A matlab toolkit for design, modeling, and control of soft robots,'' {\em IEEE Access}, vol.~PP, pp.~1--1, 01 2024.

\bibitem{Hu2019DiffTaichiDP}
Y.~Hu, L.~Anderson, T.-M. Li, Q.~Sun, N.~Carr, J.~Ragan-Kelley, and F.~Durand, ``Difftaichi: Differentiable programming for physical simulation,'' in {\em International Conference on Learning Representations}, 2020.

\bibitem{Hu2018ChainQueenAR}
Y.~Hu, J.~Liu, A.~E. Spielberg, J.~B. Tenenbaum, W.~T. Freeman, J.~Wu, D.~Rus, and W.~Matusik, ``Chainqueen: A real-time differentiable physical simulator for soft robotics,'' {\em 2019 International Conference on Robotics and Automation (ICRA)}, pp.~6265--6271, 2018.

\bibitem{articleSomogym}
M.~Graule, T.~McCarthy, C.~Teeple, J.~Werfel, and R.~Wood, ``Somogym: A toolkit for developing and evaluating controllers and reinforcement learning algorithms for soft robots,'' {\em IEEE Robotics and Automation Letters}, vol.~7, pp.~1--1, 04 2022.

\bibitem{Du2021DiffPDDP}
T.~Du, K.~Wu, P.~Ma, S.~Wah, A.~E. Spielberg, D.~Rus, and W.~Matusik, ``Diffpd: Differentiable projective dynamics,'' {\em ACM Transactions on Graphics (TOG)}, vol.~41, pp.~1 -- 21, 2021.

\bibitem{penaumticDifferentiable}
A.~Gjoka, E.~Knoop, M.~B{\"a}cher, D.~Zorin, and D.~Panozzo, ``Soft pneumatic actuator design using differentiable simulation,'' in {\em ACM SIGGRAPH 2024 Conference Papers}, pp.~1--11, 2024.

\bibitem{Huang2022DifferentiableSF}
Z.~Huang, D.~C. Tozoni, A.~Gjoka, Z.~Ferguson, T.~Schneider, D.~Panozzo, and D.~Zorin, ``Differentiable solver for time-dependent deformation problems with contact,'' {\em ACM Transactions on Graphics}, vol.~43, pp.~1 -- 30, 2022.

\bibitem{montaut2022differentiablecollisiondetectionrandomized}
L.~Montaut, Q.~Le~Lidec, A.~Bambade, V.~Petrik, J.~Sivic, and J.~Carpentier, ``Differentiable collision detection: a randomized smoothing approach,'' in {\em 2023 IEEE International Conference on Robotics and Automation (ICRA)}, pp.~3240--3246, IEEE, 2023.

\bibitem{lidec2024endtoendhighlyefficientdifferentiablesimulation}
Q.~L. Lidec, L.~Montaut, Y.~de~Mont-Marin, and J.~Carpentier, ``End-to-end and highly-efficient differentiable simulation for robotics,'' 2024.

\bibitem{Montaut2022CollisionDA}
L.~Montaut, Q.~Le~Lidec, V.~Petrik, J.~Sivic, and J.~Carpentier, ``Collision detection accelerated: An optimization perspective,'' in {\em RSS 2022-Robotics: Science and Systems (RSS)}, 2022.

\bibitem{contactLidec}
Q.~Le~Lidec, W.~Jallet, L.~Montaut, I.~Laptev, C.~Schmid, and J.~Carpentier, ``Contact models in robotics: a comparative analysis,'' {\em IEEE Transactions on Robotics}, 2024.

\bibitem{de1998bipotential}
G.~DE~SAXCE and Z.-Q. FENG, ``The bipotential method: A constructive approach to design the complete contact law with friction and improved numerical algorithms,'' {\em Mathematical and computer modelling}, vol.~28, no.~4-8, pp.~225--245, 1998.

\bibitem{acary2018solving}
V.~Acary, M.~Br{\'e}mond, and O.~Huber, ``On solving contact problems with coulomb friction: formulations and numerical comparisons,'' {\em Advanced Topics in Nonsmooth Dynamics: Transactions of the European Network for Nonsmooth Dynamics}, pp.~375--457, 2018.

\bibitem{Carpentier2024FromCT}
J.~Carpentier, Q.~Le~Lidec, and L.~Montaut, ``From compliant to rigid contact simulation: a unified and efficient approach,'' in {\em RSS 2024-Robotics: Science and Systems (RSS)}, 2024.

\bibitem{phdthesisBrunet}
J.-N. Brunet, {\em Exploring new numerical methods for the simulation of soft tissue deformations in surgery assistance}.
\newblock PhD thesis, 11 2020.

\bibitem{articleActuation}
E.~Coevoet, T.~Morales-Bieze, F.~Largilliere, Z.~Zhang, M.~Thieffry, M.~Sanz-Lopez, B.~Carrez, D.~Marchal, O.~Goury, J.~Dequidt, and C.~Duriez, ``Software toolkit for modeling, simulation, and control of soft robots,'' {\em Advanced Robotics}, vol.~31, pp.~1--17, 11 2017.

\bibitem{hppfcl}
J.~Pan, S.~Chitta, J.~Pan, D.~Manocha, J.~Mirabel, J.~Carpentier, and L.~Montaut, ``{HPP-FCL - An extension of the Flexible Collision Library},'' Mar. 2024.

\bibitem{Coevoet2017OptimizationBasedIM}
E.~Coevoet, A.~Escande, and C.~Duriez, ``Optimization-based inverse model of soft robots with contact handling,'' {\em IEEE Robotics and Automation Letters}, vol.~2, pp.~1413--1419, 2017.

\bibitem{Navarro2020AMS}
S.~E. Navarro, S.~Nagels, H.~Alagi, L.-M. Faller, O.~Goury, T.~Morales-Bieze, H.~Zangl, B.~Hein, R.~Ramakers, W.~Deferme, G.~Zheng, and C.~Duriez, ``A model-based sensor fusion approach for force and shape estimation in soft robotics,'' {\em IEEE Robotics and Automation Letters}, vol.~5, pp.~5621--5628, 2020.

\bibitem{navez2024design}
T.~Navez, B.~Li{\'e}vin, Q.~Peyron, S.~E. Navarro, O.~Goury, and C.~Duriez, ``Design optimization of a soft gripper using self-contacts,'' in {\em 2024 IEEE 7th International Conference on Soft Robotics (RoboSoft)}, pp.~1054--1060, IEEE, 2024.

\bibitem{More1977TheLA}
J.~J. Mor{\'e}, ``The levenberg-marquardt algorithm: implementation and theory,'' in {\em Numerical analysis: proceedings of the biennial Conference held at Dundee}, pp.~105--116, Springer, 1977.

\bibitem{bambade2023proxqp}
A.~Bambade, S.~El-Kazdadi, A.~Taylor, and J.~Carpentier, ``Prox-qp: Yet another quadratic programming solver for robotics and beyond,'' in {\em RSS 2022-Robotics: Science and Systems}, 2022.

\bibitem{hofer2021sim2real}
S.~Höfer, K.~Bekris, A.~Handa, J.~C. Gamboa, M.~Mozifian, F.~Golemo, C.~Atkeson, D.~Fox, K.~Goldberg, J.~Leonard, C.~Karen~Liu, J.~Peters, S.~Song, P.~Welinder, and M.~White, ``Sim2real in robotics and automation: Applications and challenges,'' {\em IEEE Transactions on Automation Science and Engineering}, vol.~18, no.~2, pp.~398--400, 2021.

\bibitem{pinocchioweb}
J.~Carpentier, F.~Valenza, N.~Mansard, {\em et~al.}, ``Pinocchio: fast forward and inverse dynamics for poly-articulated systems.'' https://stack-of-tasks.github.io/pinocchio, 2015--2024.

\bibitem{carpentier2019pinocchio}
J.~Carpentier, G.~Saurel, G.~Buondonno, J.~Mirabel, F.~Lamiraux, O.~Stasse, and N.~Mansard, ``The pinocchio c++ library -- a fast and flexible implementation of rigid body dynamics algorithms and their analytical derivatives,'' in {\em IEEE International Symposium on System Integrations (SII)}, 2019.

\end{thebibliography}
